\documentclass[letterpaper, 10 pt, conference]{ieeeconf}  

\IEEEoverridecommandlockouts                              

\usepackage{multirow}
\usepackage{tikz}
\usepackage{adjustbox}
\usepackage{algorithm}               
\usepackage{algorithmic}             
\usepackage{booktabs}
\usepackage{epstopdf}
\usepackage{enumerate}
\usepackage{pdfpages}
\usepackage{afterpage}
\usepackage{color, colortbl}
\usepackage{footmisc}
\usepackage{etoolbox}
\usepackage{tablefootnote}
\usepackage{amsmath}
\usepackage{url}
\usepackage{graphicx}

\usepackage{subfig}
\usepackage{caption}
\usepackage[flushleft]{threeparttable}
\usepackage{pifont}

\title{\LARGE \bf
FADNet: A Fast and Accurate Network for Disparity Estimation
}

\author{Qiang Wang$^{1,*}$, Shaohuai Shi$^{1,*}$, Shizhen Zheng$^{1}$, Kaiyong Zhao$^{1,\dagger}$, Xiaowen Chu$^{1,\dagger}$ 
\thanks{$^{*}$Authors have contributed equally.}
\thanks{$^{\dagger}$Corresponding authors.}
\thanks{$^{1}$Department of Computer Science, Hong Kong Baptist University, {\tt\small \{qiangwang,csshshi,szzheng,kyzhao,chxw\}\newline@comp.hkbu.edu.hk}}
}

\begin{document}

\maketitle
\thispagestyle{empty}
\pagestyle{empty}

\begin{abstract}
Deep neural networks (DNNs) have achieved great success in the area of computer vision. The disparity estimation problem tends to be addressed by DNNs which achieve much better prediction accuracy in stereo matching than traditional hand-crafted feature based methods. On one hand, however, the designed DNNs require significant memory and computation resources to accurately predict the disparity, especially for those 3D convolution based networks, which makes it difficult for deployment in real-time applications. On the other hand, existing computation-efficient networks lack expression capability in large-scale datasets so that they cannot make an accurate prediction in many scenarios. To this end, we propose an efficient and accurate deep network for disparity estimation named FADNet with three main features: 1) It exploits efficient 2D based correlation layers with stacked blocks to preserve fast computation; 2) It combines the residual structures to make the deeper model easier to learn; 3) It contains multi-scale predictions so as to exploit a multi-scale weight scheduling training technique to improve the accuracy. We conduct experiments to demonstrate the effectiveness of FADNet on two popular datasets, Scene Flow and KITTI 2015. Experimental results show that FADNet achieves state-of-the-art prediction accuracy, and runs at a significant order of magnitude faster speed than existing 3D models. The codes of FADNet are available at \url{https://github.com/HKBU-HPML/FADNet}.
\end{abstract}

\section{Introduction}
It has been seen that deep learning has been widely deployed in many computer vision tasks. Disparity estimation (also referred to as stereo matching) is a classical and important problem in computer vision applications, such as 3D scene reconstruction, robotics and autonomous driving. While traditional methods based on hand-crafted feature extraction and matching cost aggregation such as Semi-Global Matching (SGM) \cite{hirschmuller2007stereo}) tend to fail on those textureless and repetitive regions in the images, recent advanced deep neural network (DNN) techniques surpass them with decent generalization and robustness to those challenging patches, and achieve state-of-the-art performance in many public datasets \cite{zagoruyko2015learning}\cite{zbontar2016stereo}\cite{flownet}\cite{mayer2016large}\cite{psmnet}\cite{ganet}. The DNN-based methods for disparity estimation are end-to-end frameworks which take stereo images (left and right) as input to the neural network and predict the disparity directly. The architectures of DNN are very essential to achieve accurate estimation, and can be categorized into two classes, encoder-decoder network with 2D convolution (ED-Conv2D) and cost volume matching with 3D convolution (CVM-Conv3D). Besides, recent studies \cite{Saikia_2019_ICCV}\cite{he2019automl} begin to reveal the potential of automated machine learning (AutoML) for neural architecture search (NAS) on stereo matching, while some others \cite{mayer2016large}\cite{wang2019irs} focus on creating large scale datasets with high-quality labels. In practice, to measure whether a DNN model is good enough, we not only need to evaluate its accuracy on unseen samples (whether it can estimate the disparity correctly), but also its time efficiency (whether it can generate the results in real-time).
 \begin{figure}[t]
    \captionsetup[subfigure]{farskip=1pt}
	\centering
	\subfloat[Input image]
	{
	\includegraphics[width=0.45\linewidth]{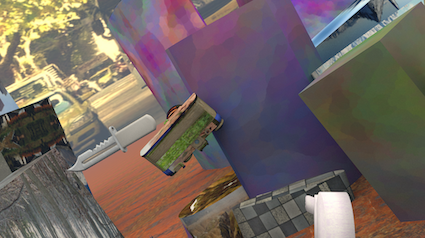}\label{fig:vt_bf_left_rgb}
	}
	\subfloat[PSMNet \cite{psmnet}]
	{
	\includegraphics[width=0.45\linewidth]{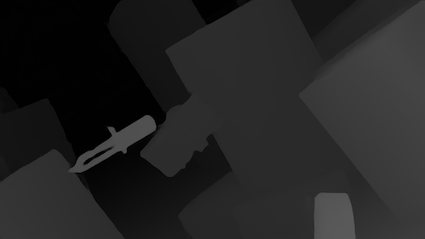}\label{fig:vt_bf_disp_gt}
	}
	\newline
	\subfloat[Our FADNet]
	{
	\includegraphics[width=0.45\linewidth]{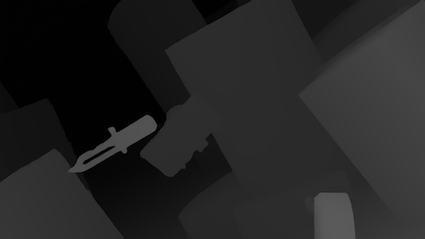}\label{fig:vt_bf_disp_irs}
	}
	\subfloat[Ground truth]
	{
	\includegraphics[width=0.45\linewidth]{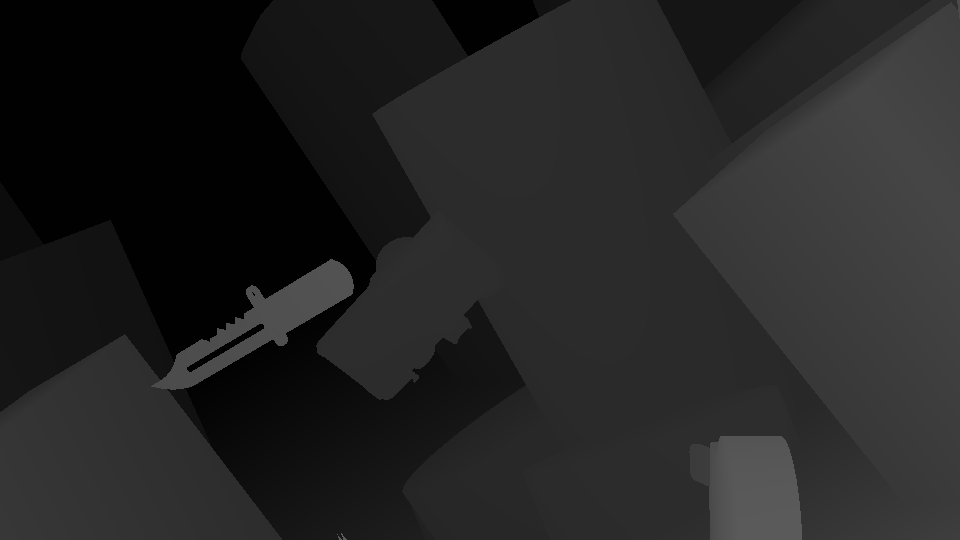}\label{fig:vt_bf_disp_sf}
	}
	\caption{Performance illustrations. (a) a challenging input image. (b) Result of PSMNet \cite{psmnet} which consumes 13.99 GB GPU memory and runs 399.3 ms for one stereo image pair on an Nvidia Tesla V100 GPU. (c) Result of our FADNet, which only consumes 1.62 GB GPU memory and runs 18.7 ms for one stereo image pair on the Nvidia Tesla V100 GPU.}
	\label{fig:results_preview}
	\vspace{-1.0 em}
\end{figure} 

In ED-Conv2D methods, stereo matching neural networks \cite{zagoruyko2015learning}\cite{zbontar2016stereo}\cite{mayer2016large} are first proposed for end-to-end disparity estimation by exploiting an encoder-decoder structure. The encoder part extracts the features from the input images, and the decoder part predicts the disparity with the generated features. The disparity prediction is optimized as a regression or classification problem using large-scale datasets (e.g., Scene Flow \cite{mayer2016large}, IRS \cite{wang2019irs}) with disparity ground truth. The correlation layer \cite{dosovitskiy2015flownet}\cite{mayer2016large} is then proposed to increase the learning capability of DNNs in disparity estimation, and it has been proved to be successful in learning strong features at multiple levels of scales \cite{dosovitskiy2015flownet}\cite{mayer2016large}\cite{flownet2}\cite{flownet3}\cite{liang2018learning}. To further improve the capability of the models, residual networks \cite{resnet}\cite{orhan2017skip}\cite{zhan2019dsnet} are introduced into those ED-Conv2D networks since the residual structure enables much deeper network to be easier to train \cite{du2019amnet}. The ED-Conv2D methods have been proved computing efficient, but they cannot achieve very high estimation accuracy.

To address the accuracy problem of disparity estimation, researchers have proposed CVM-Conv3D networks to better capture the features of stereo images and thus improve the estimation accuracy \cite{zbontar2016stereo}\cite{kendall2017end}\cite{psmnet}\cite{ganet}\cite{nie2019multi}. The key idea of the CVM-Conv3D methods is to generate the cost volume by concatenating left feature maps with their corresponding right counterparts across each disparity level \cite{kendall2017end}\cite{psmnet}. The features of cost volume are then automatically extracted by 3D convolution layers. However 3D operations in DNNs are computing-intensive and hence very slow even with current powerful AI accelerators (e.g., GPUs). Although the 3D convolution based DNNs can achieve state-of-the-art disparity estimation accuracy, they are difficult for deployment due to their resource requirements. On one hand, it requires a large amount of memory to install the model; so only a limited set of accelerators (like Nvidia Tesla V100 with 32GB memory) can run these models. On the other hand, it takes several seconds to generate a single result even on the very powerful Tesla V100 GPU using CVM-Conv3D models. The memory consumption and the inefficient computation make the CVM-Conv3D methods difficult to be deployed in practice. Therefore, it is crucial to address the accuracy and efficiency problems for real-world applications.

To this end, we propose FADNet which is a \underline{F}ast and \underline{A}ccurate \underline{D}isparity estimation \underline{Net}work based on ED-Conv2D architectures. FADNet can achieve high accuracy while keeping a fast inference speed. As illustrated in Fig. \ref{fig:results_preview}, our FADNet can easily obtain comparable performance as state-of-the-art PSMNet \cite{psmnet}, while it runs approximately 20$\times$ faster than PSMNet and consumes 10$\times$ less GPU memory. In FADNet, we first exploit the multiple stacked 2D-based convolution layers with fast computation, and then we combine state-of-the-art residual architectures to improve the learning capability, and finally we introduce multi-scale outputs for FADNet so that it can exploit the multi-scale weight scheduling to improve the training speed. These features enable FADNet to efficiently predict the disparity with high accuracy as compared to existing work. Our contributions are summarized as follows:
\begin{itemize}
    \item We propose an accurate yet efficient DNN architecture for disparity estimation named FADNet, which achieves comparable prediction accuracy as CVM-Conv3D models and it runs at an order of magnitude faster speed than the 3D-based models.
    \item We develop a multiple rounds training scheme with multi-scale weight scheduling for FADNet during training, which improves the training speed yet maintains the model accuracy.
    \item We achieve state-of-the-art accuracy on the Scene Flow dataset with up to 20$\times$ and 45$\times$ faster disparity prediction speed than PSMNet \cite{psmnet} and GANet \cite{ganet} respectively.
\end{itemize}

The rest of the paper is organized as follows. We introduce some related work in DNN based stereo matching problems in Section \ref{sec:related_work}. Section \ref{sec:model} introduces the methodology and implementation of our proposed network. We demonstrate our experimental results in Section \ref{sec:exp}. We finally conclude the paper in Section \ref{sec:conclusion}.

\begin{figure*}[htbp]
	\centering
	\includegraphics[width=0.92\linewidth]{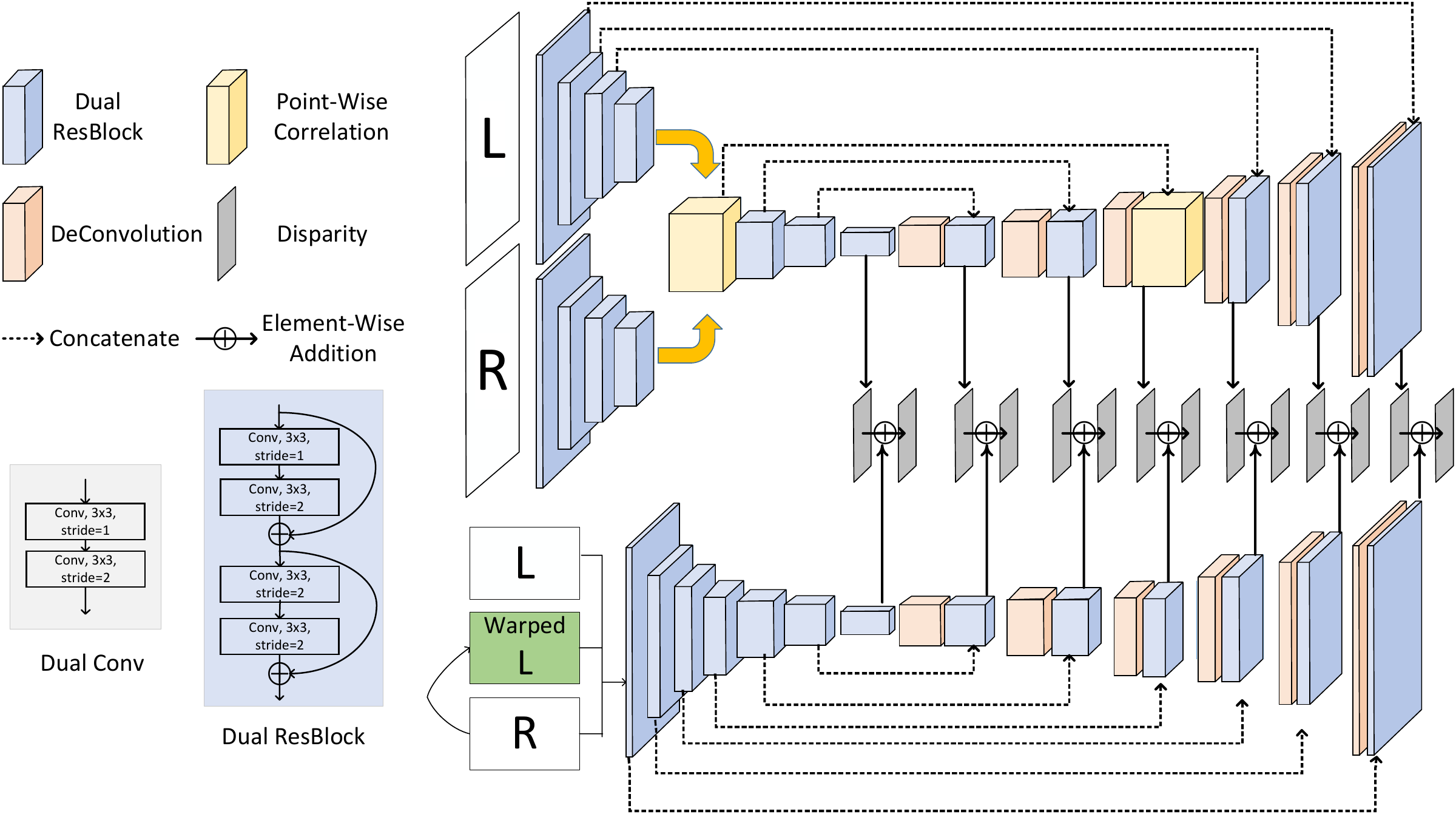}
	\caption{The model structure of our proposed FADNet.}
	\label{fig:fadnet}
	\vspace{-1.0 em}
\end{figure*}
\section{Related Work}\label{sec:related_work}
There exist many studies using deep learning methods in estimating image depth using monocular, stereo and multi-view images. Although monocular vision is low cost and commonly available in practice, it does not explicitly introduce any geometrical constraint, which is important for disparity estimation\cite{Luo_2018_CVPR}. On the contrary, stereo vision leverages the advantages of cross-reference between the left and the right view, and usually show greater performance and robustness in geometrical tasks. In this paper, we mainly discuss the work related to stereo images for disparity estimation, which is classified into two categories: 2D based and 3D based CNNs.

In 2D based CNNs, end-to-end architectures with mainly convolution layers \cite{mayer2016large}\cite{pang2017cascade} are proposed for disparity estimation, which use two stereo images as input and generate the disparity directly and the disparity is optimized as a regression task. However, the models are pure 2D CNN architectures which are difficult to capture the matching features such that the estimation results are not good. To address the problem, the correlation layer which can express the relationship between left and right images is introduced in the end-to-end architecture (e.g., DispNetCorr1D \cite{mayer2016large}, FlowNet \cite{dosovitskiy2015flownet}, FlowNet2 \cite{ilg2017flownet}, DenseMapNet \cite{atienza2018fast}). The correlation layer significantly increases the estimating performance compared to the pure CNNs, but existing architectures are still not accurate enough for production.

3D based CNNs are further proposed to increase the estimation performance \cite{zbontar2016stereo}\cite{kendall2017end}\cite{psmnet}\cite{ganet}\cite{nie2019multi}, which employ 3D convolutions with cost volume. The cost volume is mainly formed by concatenating left feature maps with their corresponding right counterparts across each disparity level \cite{kendall2017end}\cite{psmnet}, and the features of the generated cost volumes can be learned by 3D convolution layers. The 3D based CNNs can automatically learn to regularize the cost volume, which have achieved state-of-the-art accuracy of various datasets. However, the key limitation of the 3D based CNNs is their high computation resource requirements. For example, training GANet \cite{ganet} with the Scene Flow \cite{mayer2016large} dataset takes weeks even using very powerful Nvidia Tesla V100 GPUs. Even they achieve good accuracy, it is difficult to deploy due to their very low time efficiency. To this end, we propose a fast and accurate DNN model for disparity estimation.

\section{Model Design and Implementation}\label{sec:model}
Our proposed FADNet exploits the structure of DispNetC \cite{mayer2016large} as a backbone, but it is extensively reformed to take care of both accuracy and inference speed, which is lacking in existing studies. We first change the structure in terms of branch depth and layer type by introducing two new modules, residual block and point-wise correlation. Then we exploit the multi-scale residual learning strategy for training the refinement network. Finally, a loss weight training schedule is used to train the network in a coarse-to-fine manner. 

\subsection{Residual Block and Point-wise Correlation}
DispNetC and DispNetS which are both from the study in \cite{mayer2016large} basically use an encoder-decoder structure equipped with five feature extraction and down-sampling layers and five feature deconvolution layers. While conducting feature extraction and down-sampling, DispNetC and DispNetS first adopt a convolution layer with a stride of 1 and then a convolution layer with a stride of 2 so that they consistently shrink the feature map size by half. We call the two-layer convolutions with size reduction as Dual-Conv, which is shown in the left-bottom corner of Fig. \ref{fig:fadnet}. DispNetC equipped with Dual-Conv modules and a correlation layer finally achieves an end-points error (EPE) of 1.68 on the Scene Flow dataset, as reported in \cite{mayer2016large}.

The residual block originally derived in \cite{resnet} for image classification tasks is widely used to learn robust features and train a very deep networks. The residual block can well address the gradient vanish problem when training very deep networks. Thus, we replace the convolution layer in the Dual-Conv module by the residual block to construct a new module called Dual-ResBlock, which is shown in the left-bottom corner of Fig. \ref{fig:fadnet}. With Dual-ResBlock, we can make the network deeper without training difficulty as the residual block allows us to train very deep models. Therefore, we further increase the number of feature extraction and down-sampling layers from five to seven. Finally, DispNetC and DispNetS are evolving to two new networks with better learning ability, which are called RB-NetC and RB-NetS respectively, as shown in Fig. \ref{fig:fadnet}. 

One of the most important contributions of DispNetC is the correlation layer, which targets at finding correspondences between the left and right images. Given two multi-channel
feature maps $\textbf{f}_1,\textbf{f}_2$ with $w, h$ and $c$ as their width, height and number of channels, the correlation layer calculates the cost volume of them using Eq. \eqref{eq:corr}.
\begin{align}
    c(\textbf{x}_1,\textbf{x}_2)=\sum_{\textbf{o} \in [-k, k]\times [-k,k]}\langle\textbf{f}_1(\textbf{x}_1 + \textbf{o}), \textbf{f}_2(\textbf{x}_2 + \textbf{o}) \rangle, \label{eq:corr}
\end{align}
where $k$ is the kernel size of cost matching, $\textbf{x}_1$ and $\textbf{x}_2$ are the centers of two patches from $\textbf{f}_1$ and $\textbf{f}_2$ respectively. Computing all patch combinations involves $c\times K^2 \times w^2 \times h^2$ multiplication and produces a cost matching map of $w \times h$. Given a maximum searching range $D$, we fix $\textbf{x}_1$ and shift the $\textbf{x}_2$ on the x-axis direction from $-D$ to $D$ with a stride of two. Thus, the final output cost volume size will be $w\times h \times D$. 

However, the correlation operation assumes that each pixel in the patch contributes equally to the point-wise convolution results, which may loss the ability to learn more complicated matching patterns. Here we propose point-wise correlation composed of two modules. The first module is a classical convolution layer with a kernel size of $3\times3$ and a stride of $1$. The second one is an element-wise multiplication which is defined by Eq. \eqref{eq:pw_corr}.
\begin{align}
    c(\textbf{x}_1,\textbf{x}_2)=\sum{\langle\textbf{f}_1(\textbf{x}_1), \textbf{f}_2(\textbf{x}_2) \rangle}, \label{eq:pw_corr}
\end{align}
where we remove the patch convolution manner from Eq. \eqref{eq:corr}. Since the maximum valid disparity is 192 in the evaluated datasets, the maximum search range for the original image resolution is no more than 192. Remember that the correlation layer is put after the third Dual-ResBlock, of which the output feature resolution is 1/8. So a proper searching range value should not be less than 192/8=16. We set a marginally larger value 20. We also test some other values, such as 10 and 40, which do not surpass the version of using 20 in the network. The reason is that applying too small or large search range value may lead to under-fitting or over-fitting. 

Table \ref{tab:res_corr} lists the accuracy improvement brought by applying the proposed Dual-ResBlock and point-wise correlation. We train them using the same dataset as well as the training schemes. It is observed that RB-NetC outperforms DispNetC with a much lower EPE, which indicates the effectiveness of the residual structure. We also notice that setting a proper searching range value of the correlation layer helps further improve the model accuracy. 
\begin{table}[!ht]
	\centering
	\caption{Model accuracy improvement of Dual-ResBlock and point-wise correlation with different $D$.}
	\label{tab:res_corr}
		\begin{tabular}{|c|c|c|c|} \hline
			\textbf{Model} & $D$ & Training EPE & Test EPE \\ \hline\hline
			DispNetC & 20 & 2.89 & 2.80 \\ \hline
			RB-NetC &10 & 2.28 & 2.06 \\ \hline
			RB-NetC &20 & 2.09 & 1.76 \\ \hline
			RB-NetC &40 & 2.12 & 1.83 \\ \hline
		\end{tabular}
\end{table}

\subsection{Multi-Scale Residual Learning}
Instead of directly stacking DispNetC and DispNetS sub-networks to conduct disparity refinement procedure \cite{flownet3}, we apply the multi-scale residual learning firstly proposed by \cite{crl}. The basic idea is that the second refinement network learns the disparity residuals and accumulates them into the initial results generated by the first network, instead of directly predicting the whole disparity map. In this way, the second network only needs to focus on learning the highly nonlinear residual, which is effective to avoid gradient vanishing. Our final FADNet is formed by stacking RB-NetC and RB-NetS with multi-scale residual learning, which is shown in Fig. \ref{fig:fadnet}. 

As illustrated in Fig. \ref{fig:fadnet}, the upper RB-NetC takes the left and right images as input and produces disparity maps at a total of 7 scales, denoted by $c_{s}$, where $s$ is from 0 to 6. The bottom RB-NetS exploits the inputs of the left image, right image, and the warped left images to predict the residuals. The generated residuals (denoted by $r_{s}$) from RB-NetS are then accumulated to the prediction results by RB-NetC to generate the final disparity maps with multiple scales ($s=0,1,...,6$). Thus, the final disparity maps predicted by FADNet, denoted by $\hat{d_{s}}$, can be calculated by
\begin{align}
    \hat{d_{s}} = c_{s} + r_{s}, 0 \leq s \leq 6. \label{eq:d_r_sum}
\end{align}

\subsection{Loss Function Design}\label{subsec:loss}
Given a pair of stereo RGB images, our FADNet takes them as input and produces seven disparity maps at different scales. Assume that the input image size is $H \times W$. The dimension of the seven scales of the output disparity maps are $H \times W$, $\frac{1}{2}H \times \frac{1}{2}W$, $\frac{1}{4}H \times \frac{1}{4}W$, $\frac{1}{8}H \times \frac{1}{8}W$, $\frac{1}{16}H \times \frac{1}{16}W$, $\frac{1}{32}H \times \frac{1}{32}W$, and $\frac{1}{64}H \times \frac{1}{64}W$ respectively. To train FADNet in an end-to-end manner, we adopt the pixel-wise smooth L1 loss between the predicted disparity map and the ground truth using  
\begin{align}
    L_s(d_s, \hat{d_s})=\frac{1}{N}\sum_{i=1}^{N}{smooth}_{L_1}(d_{s}^i - \hat{d_{s}^i}), \label{eq:smooth_l1}
\end{align}
where $N$ is the number of pixels of the disparity map, $d_s^i$ is the $i^{th}$ element of $d_s\in \mathcal{R}^N$ and 
\begin{align}
    {smooth}_{L_1}(x)=
    \begin{cases}
    0.5x^2,& \text{if } |x| < 1\\
    |x|-0.5,              & \text{otherwise}.
\end{cases}
\end{align}
Note that $d_s$ is the ground truth disparity of scale $\frac{1}{2^s}$ and $\hat{d_s}$ is the predicted disparity of scale $\frac{1}{2^s}$. The loss function is separately applied in the seven scales of outputs, which generates seven loss values. The loss values are then accumulated with loss weights. 

\begin{table}[!ht]
	\centering
	\caption{Multi-scale loss weight scheduling.}
	\label{tab:loss_weights}
		\begin{tabular}{|c|c|c|c|c|c|c|c|} \hline
		\textbf{Round}	& $w_0$ & $w_1$ & $w_2$ & $w_3$ & $w_4$ & $w_5$ & $w_6$ \\ \hline\hline
			1 & 0.32 & 0.16 & 0.08 & 0.04 & 0.02 & 0.01 & 0.005 \\ \hline
			2 & 0.6 & 0.32 & 0.08 & 0.04 & 0.02 & 0.01 & 0.005 \\ \hline
			3 & 0.8 & 0.16 & 0.04 & 0.02 & 0.01 & 0.005 & 0.0025 \\ \hline
			4 & 1.0 & 0 & 0 & 0 & 0 & 0 & 0\\ \hline
		\end{tabular}
\end{table}

The loss weight scheduling technique which is initially proposed in \cite{mayer2016large} is useful to learn the disparity in a coarse-to-fine manner. Instead of just switching on/off the losses of different scales, we apply different non-zero weight groups for tackling different scale of disparity. Let $w_s$ denote the weight for the loss of the scale of $s$. The final loss function is
\begin{equation}
    L=\sum_{s=0}^{6}w_sL_s(d_s,\hat{d_s}).
\end{equation}
The specific setting is listed in Table \ref{tab:loss_weights}. Totally there are seven scales of predicted disparity maps. At the beginning, we assign low-value weights for those large scale disparity maps to learn the coarse features. Then we increase the loss weights of large scales to let the network gradually learn the finer features. Finally, we deactivate all the losses except the final predict one of the original input size. With different rounds of weight scheduling, the evaluation EPE is gradually increased to the final accurate performance which is shown in Table \ref{tab:weightsresults} on the Scene Flow dataset.

\begin{table}[!ht]
	\centering
	\caption{Model accuracy with different rounds of weight scheduling.}
	\label{tab:weightsresults}
		\begin{tabular}{|c|c|c|c|c|} \hline
		    Round & \# Epochs & Training EPE & Test EPE & Improvement (\%) \\ \hline\hline
			1 & 20 & 1.85 & 1.57 & - \\ \hline
			2 & 20 & 1.33 & 1.32 & 18.9 \\ \hline
			3 & 20 & 1.04 & 0.93 & 41.9 \\ \hline
			4 & 30 & 0.92 & 0.83 & 12.0 \\ \hline
		\end{tabular}
		\begin{tablenotes}
    	    \item Note: ``Improvement'' indicates the improvement of the current round of weight schedule over its previous.
        \end{tablenotes}
\end{table}
Table \ref{tab:weightsresults} lists the model accuracy improvements (around 12\%-41\%) brought by the multiple round training of four loss weight groups. It is observed that both the training and testing EPEs are decreased smoothly and close, which indicates good generalization and advantages of our training strategy. 

\section{Performance Evaluation}\label{sec:exp}
In this section, we present the experimental results of our proposed FADNet compared to existing work (i.e., DispNetC \cite{mayer2016large}, PSMNet \cite{psmnet}, GANet \cite{ganet} and DenseMapNet \cite{atienza2018fast}) in terms of accuracy and time efficiency. 

\subsection{Experimental Setup}
We implement our FADNet using PyTorch\footnote{\url{https://pytorch.org}}, which is one of popular deep learning frameworks, and we make the codes and experimental setups be publicly available\footnote{\url{https://github.com/HKBU-HPML/FADNet}}.

In terms of accuracy, the model is trained with Adam (${\beta}_1=0.9, {\beta}_2=0.999$). We perform color normalization with the mean ([0.485, 0.456, 0.406]) and variation ([0.229, 0.224, 0.225]) of the ImageNet \cite{deng2009imagenet} dataset for data pre-processing. During training, images are randomly cropped to size $H=384$ and $W=768$. The batch size is set to 16 for the training on four Nvidia Titan X (Pascal) GPUs (each of 4). We apply a four-round training scheme illustrated in Section \ref{subsec:loss}, where each round adopts one different loss weight group. At the beginning of each round, the learning rate is initialized as $10^{-4}$ and is decayed by half every 10 epochs. We train 20 epochs for the first three rounds and 30 for the last round. 

In terms of time efficiency, we evaluate the inference time of existing state-of-the-art DNNs including both 2D and 3D based networks using a pair of stereo images ($H=576, W=960$) from the Scene Flow dataset \cite{mayer2016large} on a desktop-level Nvidia Titan X (Pascal) GPU (with 12GB memory) and a server-level Nvidia Tesla V100 GPU (with 32GB memory). 

\subsection{Dataset}
We used two publicly popular available datasets to train and evaluate the performance of our FADNet. The first one is Scene Flow which is produced by synthetic rendering techniques. The second one is KITTI 2015 which is captured by real world cameras and laser sensors. 
\subsubsection{Scene Flow \cite{mayer2016large}} a large synthetic dataset which provides totally 39,824 samples of stereo RGB images (35,454 for training and 4,370 for testing). The full resolution of the images is 960$\times$540. The dataset covers a wide range of object shapes and texture and provides high-quality dense disparity ground truth. We use the endpoint error (EPE) as error measurement. We remove those pixels whose disparity values are larger than 192 in the loss computation, which is typically done by the previous studies \cite{psmnet}\cite{ganet}. 
\subsubsection{KITTI 2015 \cite{Menze2015ISA}} an open benchmark dataset which contains 200 stereo images which are grayscale and have a resolution of 1241$\times$376. The ground truth of disparity is generated by the LIDAR equipment, so the disparity map is very sparse. During training, we randomly crop 1024$\times$256 resolution of images and disparity maps. We use its full resolution during test.

\subsection{Experimental Results}
\begin{table}[!t]
	\centering
	\caption{Disparity EPE on the scene flow dataset}
	\label{tab:expresults}
		\begin{tabular}{|c|c|c|c|c|} \hline
			\multirow{2}{*}{\textbf{Model}} & \multirow{2}{*}{\textbf{EPE}} & \textbf{Memory} & \multicolumn{2}{c|}{\textbf{Runtime (ms)}} \\ \cline{4-5}
			&  & \textbf{(GB)} & \textbf{Titan X (Pascal)} & \textbf{Tesla V100} \\\hline
			FADNet(ours) & \textbf{0.83} & 3.87 & 65.5 & 48.1 \\ \cline{1-5}
			DispNetC & 1.68 & 1.62 & 28.7 & 18.7 \\ \cline{1-5}
		    DenseMapNet & \underline{5.36} & - & \underline{$<$30} & - \\ \cline{1-5}
			PSMNet & 1.09 & 13.99 & OOM & 399.3 \\ \cline{1-5}
			GANet & \underline{0.84} & 29.1 & OOM & 2251.1 \\ \cline{1-5}
		\end{tabular}
		\begin{tablenotes}
    	    \item Note: ``OOM'' indicates that it runs out of memory. Runtime is the inference time per pair of stereo images, and it is measured by 100 runs with average. The underline numbers are from the original paper.
        \end{tablenotes}
\end{table}

 \begin{figure*}[ht]
    \captionsetup[subfigure]{labelformat=empty, farskip=0pt}
	\centering
	\subfloat[]
	{
	\includegraphics[width=0.235\linewidth]{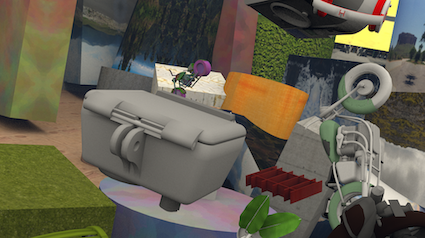}\label{fig:vt_bf_left_rgb}
	}
	\subfloat[]
	{
	\includegraphics[width=0.235\linewidth]{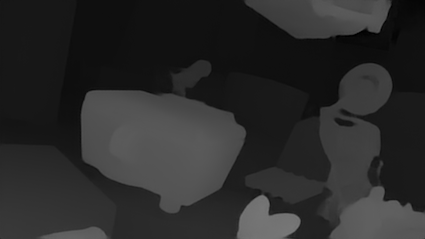}\label{fig:vt_bf_disp_gt}
	}
	\subfloat[]
	{
	\includegraphics[width=0.235\linewidth]{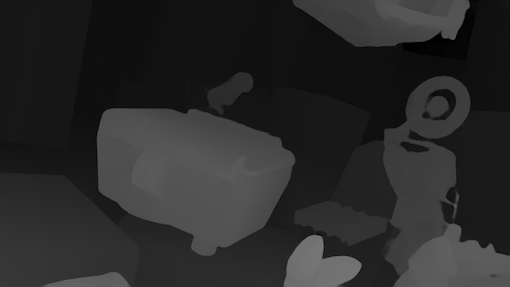}\label{fig:vt_bf_disp_irs}
	}
	\subfloat[]
	{
	\includegraphics[width=0.235\linewidth]{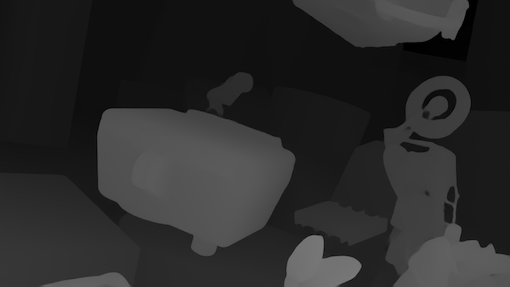}\label{fig:vt_bf_disp_sf}
	}
	\newline
 	\vspace{-1.0 em}
	\subfloat[]
	{
	\includegraphics[width=0.235\linewidth]{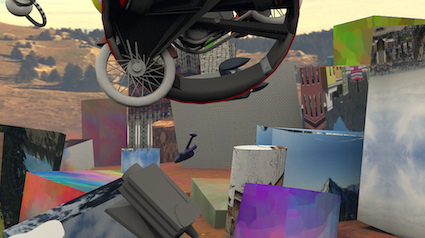}\label{fig:vt_bf_left_rgb}
	}
	\subfloat[]
	{
	\includegraphics[width=0.235\linewidth]{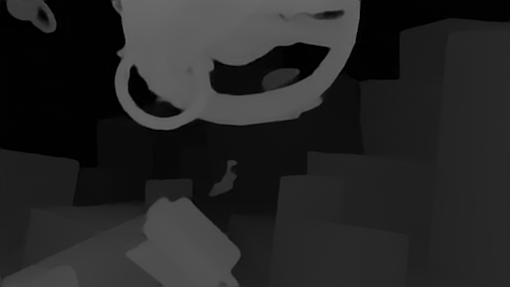}\label{fig:vt_bf_disp_gt}
	}
	\subfloat[]
	{
	\includegraphics[width=0.235\linewidth]{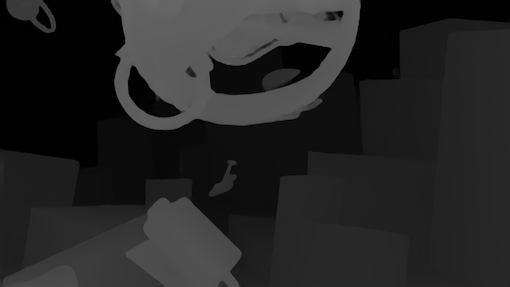}\label{fig:vt_bf_disp_irs}
	}
	\subfloat[]
	{
	\includegraphics[width=0.235\linewidth]{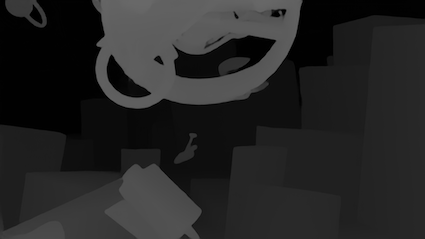}\label{fig:vt_bf_disp_sf}
	}
	\newline
 	\vspace{-1.0 em}
	\subfloat[]
	{
	\includegraphics[width=0.235\linewidth]{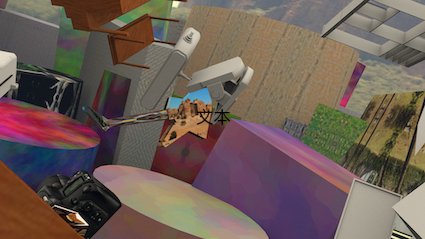}\label{fig:vt_bf_left_rgb}
	}
	\subfloat[(a) DispNetC]
	{
	\includegraphics[width=0.235\linewidth]{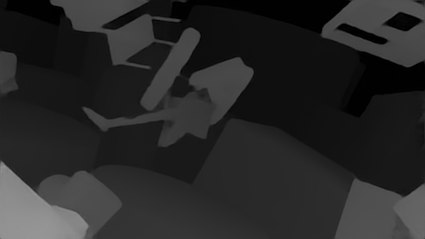}\label{fig:vt_bf_disp_gt}
	}
	\subfloat[(b) PSMNet]
	{
	\includegraphics[width=0.235\linewidth]{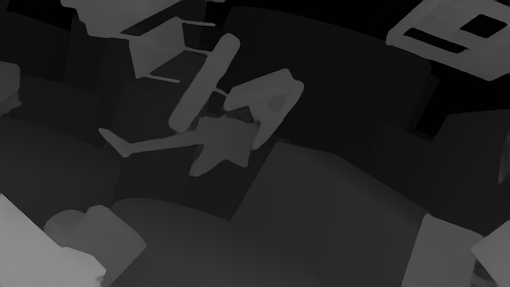}\label{fig:vt_bf_disp_irs}
	}
	\subfloat[(c) FADNet]
	{
	\includegraphics[width=0.235\linewidth]{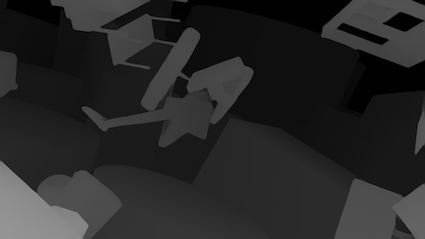}\label{fig:vt_bf_disp_sf}
	}
	\caption{Results of disparity prediction for Scene Flow testing data. The leftest column shows the left images of the stereo pairs. The rest three columns respectively show the disparity maps estimated by (a) DispNetC \cite{mayer2016large}, (b) PSMNet \cite{psmnet}, (c) FADNet. }
	\label{fig:results_on_flying}
\end{figure*} 
\begin{figure*}[ht]
    \captionsetup[subfigure]{labelformat=empty, farskip=0pt}
	\centering
	\subfloat[]
	{
	\includegraphics[width=0.235\linewidth]{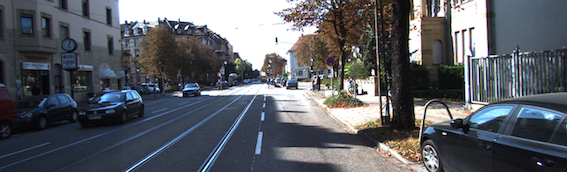}\label{fig:vt_bf_left_rgb}
	}
	\subfloat[]
	{
	\includegraphics[width=0.235\linewidth]{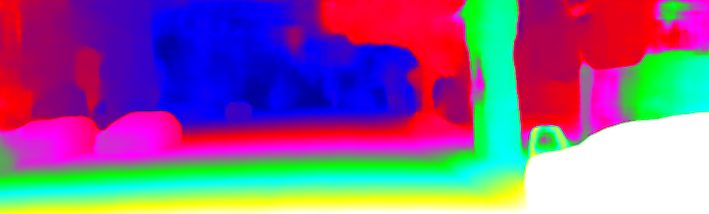}\label{fig:vt_bf_disp_gt}
	}
	\subfloat[]
	{
	\includegraphics[width=0.235\linewidth]{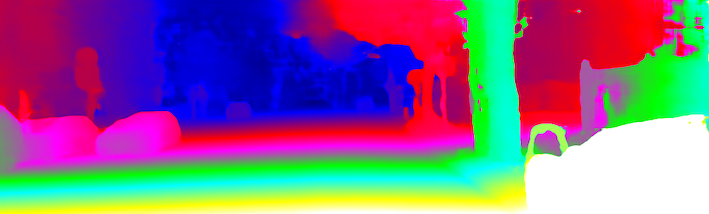}\label{fig:vt_bf_disp_irs}
	}
	\subfloat[]
	{
	\includegraphics[width=0.235\linewidth]{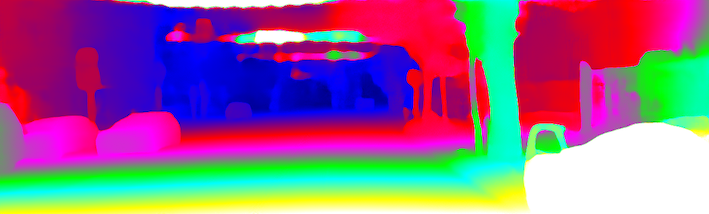}\label{fig:vt_bf_disp_sf}
	}
	\newline
 	\vspace{-1.0 em}
	\subfloat[]
	{
	\includegraphics[width=0.235\linewidth]{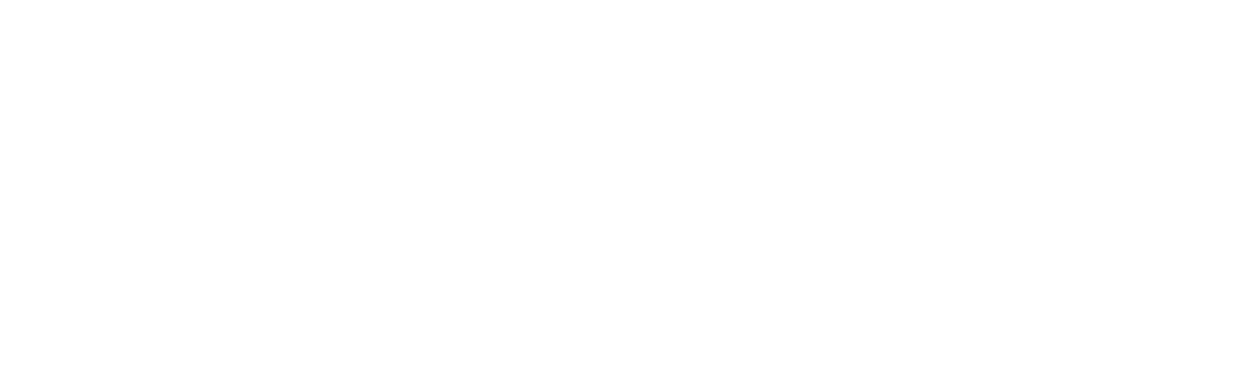}\label{fig:vt_bf_disp_sf}
	}
	\subfloat[]
	{
	\includegraphics[width=0.235\linewidth]{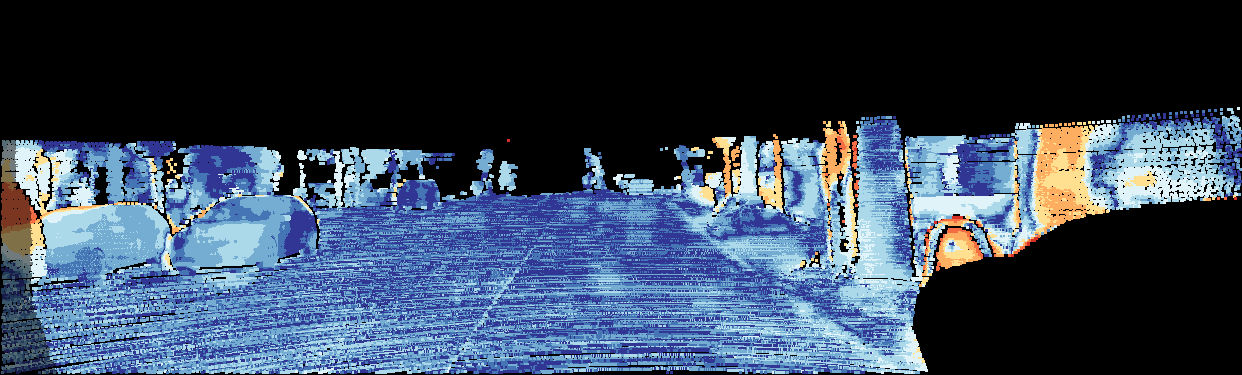}\label{fig:vt_bf_disp_gt}
	}
	\subfloat[]
	{
	\includegraphics[width=0.235\linewidth]{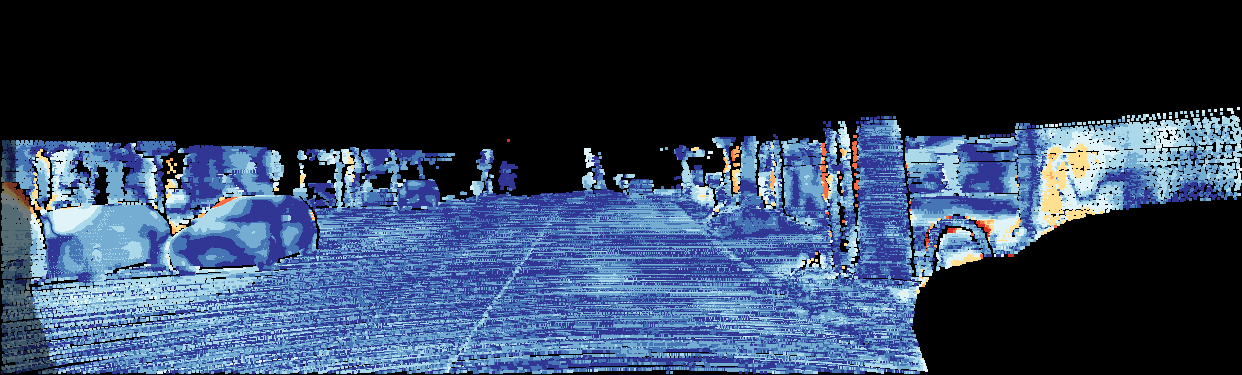}\label{fig:vt_bf_disp_irs}
	}
	\subfloat[]
	{
	\includegraphics[width=0.235\linewidth]{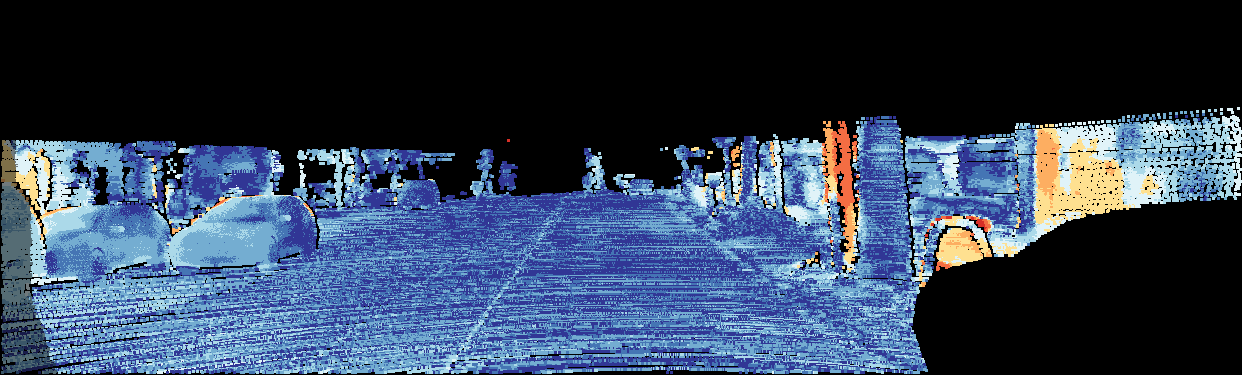}\label{fig:vt_bf_disp_sf}
	}
	\newline
 	\vspace{-1.0 em}
	\subfloat[]
	{
	\includegraphics[width=0.235\linewidth]{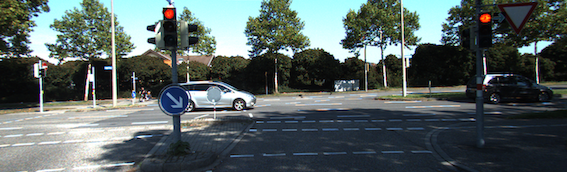}\label{fig:vt_bf_left_rgb}
	}
	\subfloat[]
	{
	\includegraphics[width=0.235\linewidth]{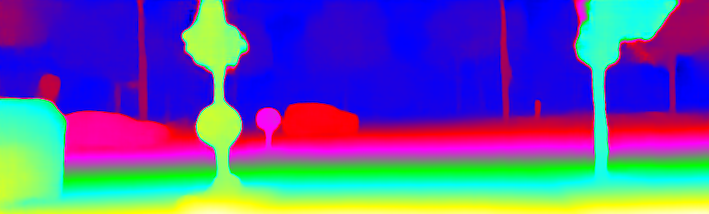}\label{fig:vt_bf_disp_gt}
	}
	\subfloat[]
	{
	\includegraphics[width=0.235\linewidth]{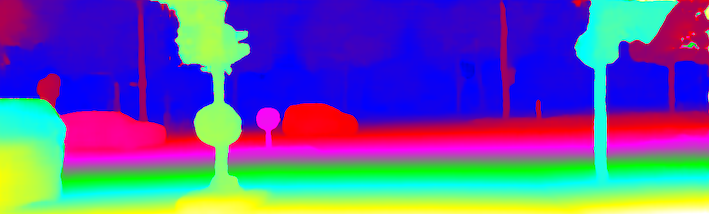}\label{fig:vt_bf_disp_irs}
	}
	\subfloat[]
	{
	\includegraphics[width=0.235\linewidth]{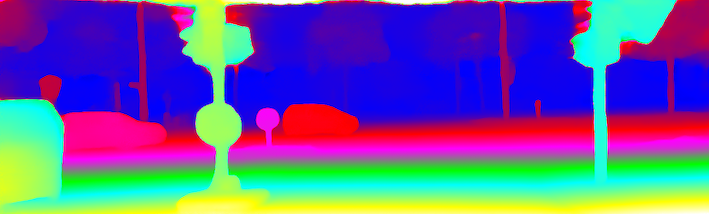}\label{fig:vt_bf_disp_sf}
	}
	\newline
 	\vspace{-1.0 em}
	\subfloat[]
	{
	\includegraphics[width=0.235\linewidth]{Figures/KITTI/blank.png}\label{fig:vt_bf_disp_sf}
	}
	\subfloat[(a) DispNetC]
	{
	\includegraphics[width=0.235\linewidth]{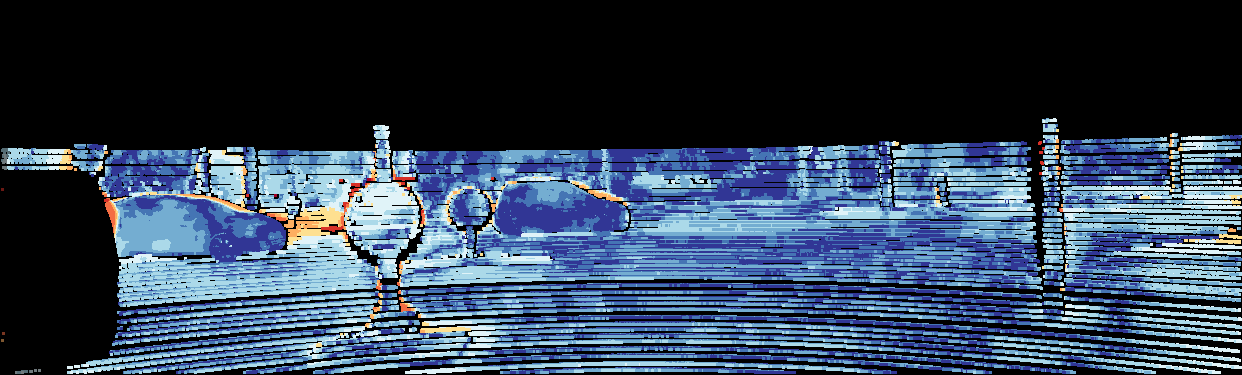}\label{fig:vt_bf_disp_gt}
	}
	\subfloat[(b) PSMNet]
	{
	\includegraphics[width=0.235\linewidth]{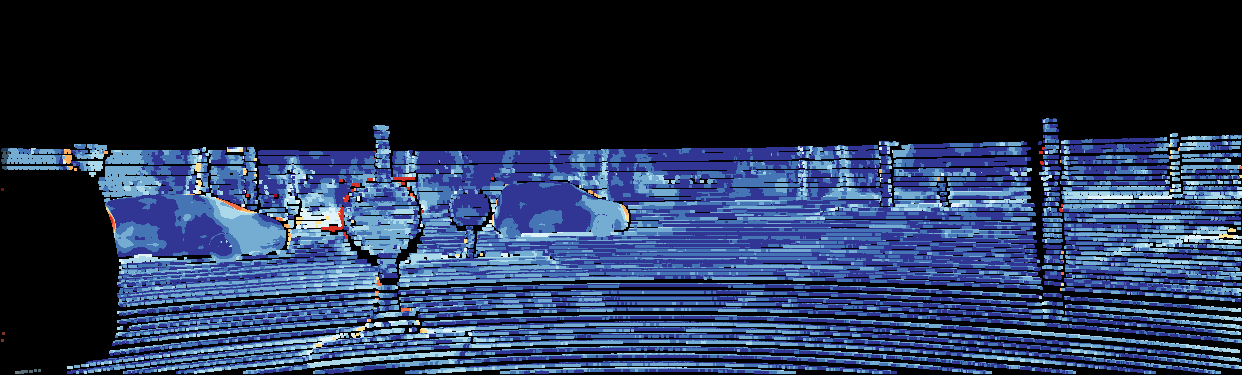}\label{fig:vt_bf_disp_irs}
	}
	\subfloat[(c) FADNet]
	{
	\includegraphics[width=0.235\linewidth]{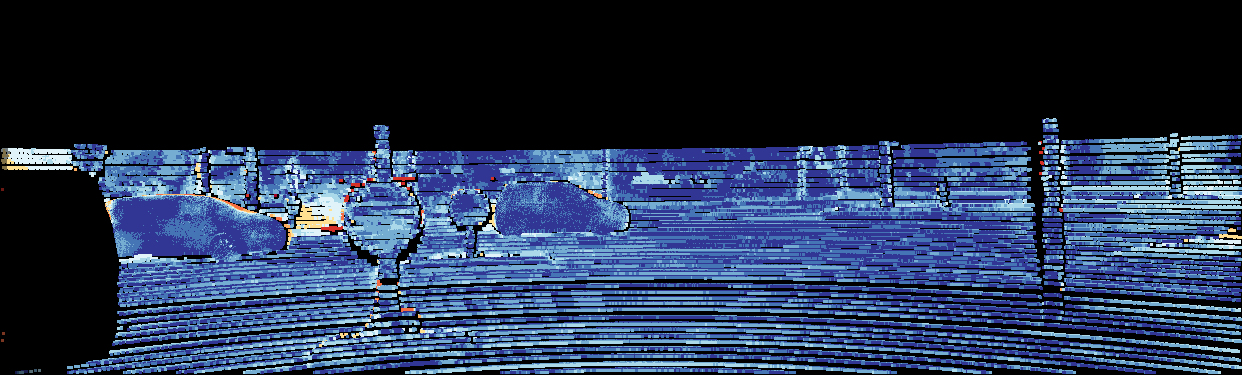}\label{fig:vt_bf_disp_sf}
	}
	\caption{Results of disparity prediction for KITTI 2015 testing data. The leftest column shows the left images of the stereo pairs. The rest three columns respectively show the disparity maps estimated by (a) DispNetC \cite{mayer2016large}, (b) PSMNet \cite{psmnet}, (c) FADNet, as well as their error maps. }
	\label{fig:results_on_kitti}
	\vspace{-0.4 em}
\end{figure*}

The experimental results on the Scene Flow dataset are shown in Table \ref{tab:expresults}. Regarding the model accuracy measured with EPE, our proposed FADNet achieves comparable performance compared to the state-of-the-art CVM-Conv3D (PSMNet and GANet), while FADNet is 46$\times$ and 8$\times$ faster than GANet and PSMNet respectively on an Nvidia Tesla V100 GPU. Even PSMNet and GANet are not runnable on the Titan X (Pascal) GPU, which implies high cost of them in practice. Compared to DispNetC and DenseMapNet, even FADNet is relatively slow, it predicts the disparity more than 2$\times$ accurate than DispNetC and DenseMapNet, which is a huge accuracy improvement. The visualized comparison with predicted disparity maps are shown in Fig. \ref{fig:results_on_flying}. 

 \begin{table}[ht]
	\centering
	\caption{Results on the KITTI 2015 dataset}
	\label{tab:perfkitti}
	\scriptsize{
		\begin{tabular}{|c|c|c|c|c|c|c|} \hline
			\textbf{Model} & \multicolumn{3}{c|}{Noc(\%)} & \multicolumn{3}{c|}{All(\%)} \\ \cline{2-7}
			& D1-bg & D1-fg & D1-all & D1-bg & D1-fg & D1-all \\ \cline{1-7}
			FADNet(ours) & 2.49\% & \textbf{3.07\%} & 2.59\% & 2.68\% & 3.50\% & 2.82\% \\ \hline
			DispNetC & 4.11\% & 3.72\% & 4.05\% & 4.32\% & 4.41\% & 4.34\% \\ \hline
			GC-Net & 2.02\% & 5.58\% & 2.61\% & 2.21\% & 6.16\% & 2.87\% \\ \hline
			PSMNet & 1.71\% & 4.31\% & 2.14\% & 1.86\% & 4.62\% & 2.32\% \\ \hline
			GANet & \textbf{1.34\%} & 3.11\% & \textbf{1.63\%} & \textbf{1.48\%} & \textbf{3.46\%} & \textbf{1.81\%} \\ \hline
		\end{tabular}
			}
		\begin{tablenotes}
    	    \item Note: ``Noc'' and ``All'' indicates percentage of outliers averaged over ground truth pixels of non-occluded and all regions respectively. ``D1-bg'', ``D1-fg'' and ``D1-all'' indicates percentage of outliers averaged over background, foreground and all ground truth pixels respectively.
        \end{tablenotes}
	
\end{table}

From the visualized disparity maps shown in Fig. \ref{fig:results_on_flying}, we can see that the details of textures are successfully estimated by our FADNet while PSMNet is a little worse and DispNetC almost misses all the details. The visualization results are dramatically different although the EPE gap between DispNetC and FADNet is only 0.85. In the qualitative evaluation, FADNet is more robust and accurate than DispNetC with a 2D based network and PSMNet with a 3D based network.

From Table \ref{tab:expresults}, it is noticed that the CVM-Conv3D architectures cannot be used on the desktop-level GPU which is equipped with 12 GB memory, while the proposed FADNet requires only 3.87 GB to perform the disparity estimation. The low memory requirement of FADNet makes it much easier for deployment in real-world applications. DispNetC is also an efficient architecture in terms of both memory consumption and computing efficiency, but its estimation performance is bad such that it cannot be used in real-world applications. In summary, FADNet not only achieves high disparity estimation accuracy, but it is also very efficient and practical for deployment.

The experimental results on the KITTI 2015 dataset are shown in Table \ref{tab:perfkitti}. GANet achieves the best estimation results among the evaluated models, and our proposed FADNet performs comparable error rates on the metric of D1-fg. The qualitative evaluation of the KITTI 2015 dataset is shown in Fig. \ref{fig:results_on_kitti}, it is seen that the error maps of FADNet are close to PSMNet, while they are much better than that of DispNetC.

\section{Conclusion and Future Work}\label{sec:conclusion}
In this paper, we proposed an efficient yet accurate neural network, FADNet, for end-to-end disparity estimation to embrace the time efficiency and estimation accuracy on the stereo matching problem. The proposed FADNet exploits point-wise correlation layers, residual blocks, and multi-scale residual learning strategy to make the model be accurate in many scenarios while preserving fast inference time. We compared FADNet with existing state-of-the-art 2D and 3D based methods on two popular datasets in terms of accuracy and speed. Experimental results showed that FADNet achieves comparable accuracy while it runs much faster than the 3D based models. Compared to the 2D based models, FADNet is more than two times accurate.

We have two future directions following our discovery in this paper. First, we would like to develop fast disparity inference of FADNet on edge devices. Since the computational capability of them is much lower than that of the server GPUs used in our experiments, it is necessary to explore the techniques of model compression, including pruning, quantization, and so on. Second, we would also like to apply AutoML \cite{he2019automl} for searching a well-performing network structure for disparity estimation.

\section*{Acknowledgements}
This research was supported by Hong Kong RGC GRF grant HKBU 12200418. We thank the anonymous reviewers for their constructive comments and suggestions. We would also like to thank NVIDIA AI Technology Centre (NVAITC) for providing the GPU clusters for some experiments.

\bibliographystyle{IEEEtran}
\bibliography{root.bbl}

\end{document}